# Diversification Methods for Zero-One Optimization

Fred Glover
ECEE- College of Engineering and Applied Science
University of Colorado - Boulder
Boulder, CO, 80309 USA
glover@colorado.edu

## Abstract

We introduce new diversification methods for zero-one optimization that significantly extend strategies previously introduced in the setting of metaheuristic search. Our methods incorporate easily implemented strategies for partitioning assignments of values to variables, accompanied by processes called augmentation and shifting which create greater flexibility and generality. We then show how the resulting collection of diversified solutions can be further diversified by means of permutation mappings, which equally can be used to generate diversified collections of permutations for applications such as scheduling and routing. These methods can be applied to non-binary vectors by the use of binarization procedures and by Diversification-Based Learning (DBL) procedures which also provide connections to applications in clustering and machine learning. Detailed pseudocode and numerical illustrations are provided to show the operation of our methods and the collections of solutions they create.

# 1. Introduction

Diversification strategies are now widely recognized as a critical part of effective metaheuristics for complex optimization problems. The important class of zero-one optimization problems is especially relevant for designing diversification strategies, because of the wide range of applications in which they arise. In addition, many discrete optimization problems can be conveniently translated into zero-one problems or can be treated using neighborhood spaces equivalent to those of zero-one problems through the design of metaheuristic search methods.

Diversification for zero-one optimization can also be applied to nonlinear continuous (global) optimization, taking advantage of the fact that binarization methods developed for converting discrete and continuous data into binary data (Mayoraz and Moreira, 1999) have proved to be quite effective for making certain types of global continuous problems susceptible to solution by zero-one optimization, notably in the realms of cluster analysis and machine learning.

Diversification is treated here in the sense proposed in adaptive memory programming (tabu search), where the drive to obtain diverse new solutions goes hand-in-hand with intensification processes, which concentrate the search more strongly in regions anticipated to contain good solutions. Consequently, our prescriptions are assumed to operate within contexts where restrictions are imposed on the search space, as in assigning bounds or fixed values to particular variables (e.g., in exploiting strongly determined and consistent variables; as in Glover (1977, 2001) and Glover and Laguna (1997)).

In this paper we introduce new diversification strategies for zero-one optimization that extend a framework for generating diverse collections of zero-one vectors originally proposed in the context of the Scatter Search and Path Relinking evolutionary algorithms (Glover, 1997). The two principal diversification strategies from this source constitute a *Progressive Gap* method and a *Max/Min* method. The Progressive Gap method has been incorporated in several studies for applying evolutionary metaheuristics to zero-one optimization problems (see, e.g., Laguna an Marti, 2003), while the Max/Min method has advantages for achieving certain kinds of diversification, and is relevant to the topic of learning procedures for metaheuristic optimization, as embodied in the approach called Diversification-based Learning (DBL) (Glover and Hao, 2017). Further connections with learning strategies derive from the fact that DBL includes methods for basing the treatment of general vectors on the ability to handle zero-one vectors.

We begin by taking ideas underlying the Max/Min method as a starting point to provide new and more advanced methods for generating diversified collections of zero-one vectors, showing how to partition the space of solutions in more refined ways to create diverse collections. Building on this, we then give an Augmented-Max/Min generation method that provides greater flexibility for creating diversified collections, and identify an associated Shifting Procedure that extends the scope of these methods. Finally, we introduce permutation mappings that further enlarge the range of diversified solutions produced, yielding solutions with new structures through a recursive application of these mappings. Our methods are accompanied by numerical examples that illustrate their operation and the collections of solutions they create.

## 1.1 Basic Notation and Conventions

In the following, the 0-1 vectors generated are denoted by x(r), for r = 0 to rLast, where x(0) denotes the seed vector x = ($x_1$, $x_2$, ..., $x_n$). The seed vector can be provided by the user, and in the case of binary optimization, can be selected to be a locally optimal 0-1 solution or derived from a linear combination of such local optima for the problem of interest.

For a given 0-1 vector x', Comp(x') denotes the complemented vector x" given by $x_j$" = 1 − $x_j$',
    j = 1, ..., n.
⌊v⌋ denotes the integer floor function which identifies the largest integer ≤ v, for any real value v.
    (Consequently, ⌊v + .5⌋ is the nearest integer neighbor of v.)
rLim is a user-selected upper limit on rLast, the number of vectors in the collection x(r), for
    r = 0 to rLast.

Each point x' or x" generated is a shorthand for identifying a current point x(r). Hence, when an algorithm assigns a particular value $x_j$' or $x_j$", it is understood that $x_j$(r) ≡ $x_j$' or $x_j$(r) ≡ $x_j$". In instances where $x_j$' and $x_j$" are determined together, it is understood that $x_j$' refers to x(r) and $x_j$" refers to x(r+1).

Several parts of this paper deal with the challenge of increasing the number of vectors to be included in a diversified collection. It should be noted that increasing the number of vectors generated does not in itself increase the diversity of the collection, or more precisely, the Mean Diversity measured by the value Mean(|x − y|: for all pairs (x,y) in the collection). For example, the greatest Mean Diversity results for a collection of just two points, consisting of a complementary pair (x', x"). Adding any additional point y compels the distance |x' − y| and |x" − y| to be less than |x' − x"|, and in general, if a set of points has been generated with a maximal Mean Diversity, adding more points will not increase the diversity by this measure.

In general, the smaller the number of points that are generated, the greater the (mean) diversity that can be achieved. However, a larger number of points can increase a different type of diversity, which involves the "coverage" provided by the points selected. (For example, one may define coverage = Mean Diversity/Mean Gap, where Gap(x, y) = |x − y| restricted to pairs (x, y) such that there is no point z closer to x than y or closer to y than x on the line segment joining x and y.) Although we do not attempt here to provide formal relationships joining these notions, it should be clear that adding more points can indeed improve the coverage. Hence, it is useful to select the limit rLim to be as large as reasonably possible, taking into account the computational tradeoffs of working with a larger number of points, as determined by the method that utilizes these points. An advantage of generating additional vectors is that it helps to combine intensification with diversification when selecting a best vector from the resulting set.
The tradeoffs between relative diversity and the number of vectors produced is a recurring theme throughout the remainder of this paper.

## 2. The Max/Min 0-1 Diversification Method

The strategy underlying the Max/Min algorithm, which we examine in several variations throughout subsequent sections, is to successively partition the indexes of x into equal sized subsets, so that each vector x' and its complement x" in the resulting sequence is separated from previous vectors by maximizing the minimum Hamming distance to these vectors. (This property can be achieved strictly when n is a power of 2, and can be achieved approximately for other values of n.)

The criterion of maximizing the minimum distance between vectors in the collection generated rests on the following observation. If x' differs from x in half of its components, this implies that the complement x" of x' will likewise differ from x in half of its components, and consequently the minimum distance of x' and x" to x will be maximized. This same criterion also implies that x' and x" will be (approximately) equidistant from the vector Comp(x), and hence the property of maximizing the minimum separating distance will hold in relation to Comp(x) as well.

### 2.1 Overview

Let $N(i)$, $i = 1, \ldots,$ iLast denote a partition of $N = \{1, \ldots, n\}$. At each stage of the method, each set $N(i)$ of the current partition is split into two equal parts (or as nearly as possible when $N(i)$ contains an odd number of elements), creating a total of iLast additional sets $N(i)$. Let $\lceil v \rceil$ denote the integer ceiling of v, i.e., the least integer $\geq v$ (hence $\lceil v \rceil = \lfloor v \rfloor + 1$ if v is fractional).

To begin, iLast = 1 and $N(1) = N = \{1, \ldots, n\}$. $N(1)$ is then split into left and right "halves" $N_L(1)$ and $N_R(1)$ so that the first $n' = \lceil n/2 \rceil$ of $N(1)$'s elements go in $N_L(1)$ and the remaining $n - n'$ elements go in $N_R(1)$, i.e., $N_L(1) = \{1, \ldots, n'\}$ and $N_R(1) = \{n'+1, \ldots, n\}$. At the conclusion of this split we update the partition by setting $N(1) = N_L(1)$ and $N(2) = N_R(1)$, thus doubling the number iLast of current sets in the partition to become 2.

On the next iteration, each of $N(1)$ and $N(2)$ are similarly split, generating sets $N_L(1)$ and $N_R(1)$ from $N(1)$ and $N_L(2)$ and $N_R(2)$ from $N(2)$. Then the updated partition is created by redefining $N(1) = N_L(1)$, $N(2) = N_R(1)$, $N(3) = N_L(2)$ and $N(4) = N_R(2)$, and thus yielding iLast = 4. In general, each time the sets in the partition $N(i)$, $i = 1$ to iLast are split, each set $N(i)$ is subdivided by the following *alternating assignment* rule. If i is odd, the first $\lceil |N(i)|/2 \rceil$ elements of $N(i)$ go into $N_L(i)$, while if i is even, the first $\lfloor |N(i)|/2 \rfloor$ elements of $N(i)$ go into $N_L(i)$. In each instance, remaining elements of $N(i)$ go into $N_R(i)$.

As each set $N(i)$ is split, before doubling iLast, we generate a new vector x' from x by setting

$$x_j' = 1 - x_j \text{ for } j \in N_L(i), \ i = 1, \ldots, \text{iLast}, \qquad (2.1)$$
$$x_j' = x_j \text{ for } j \in N_R(i), \ i = 1, \ldots, \text{iLast}. \qquad (2.2)$$

We also generate a second vector x" = Comp(x'), or equivalently,

$$x_j'' = x_j \text{ for } j \in N_L(i), \ i = 1, \ldots, \text{iLast}, \qquad (2.3)$$

$$x_j'' = 1 - x_j \text{ for } j \in N_R(i), \ i = 1, \ldots, \text{iLast}. \tag{2.4}$$

Finally the partition is updated by defining $N(2i - 1) = N_L(i)$ and $N(2i) = N_R(i)$ for $i = 1$ to iLast, followed by doubling iLast.

Each partition created by the "odd/even" rule for splitting the sets satisfies the property that $|N(i)| =$ MaxNum or $|N(i)| = $ MaxNum $- 1$, where MaxNum = Max($N(i)$, $i = 1, \ldots,$ iLast). Moreover, the organization of the method also assures MaxNum = $|N(1)|$. After some number of partitions have been generated, the value of MaxNum for the current partition will equal 2. (MaxNum may skip over some values as it successively decreases in the creation of new partitions, but will not skip over the value 2.) Once MaxNum = 2, the concluding step of the algorithm operates as follows. We identify the number Num2 of sets $N(i)$ for $i = 1$ to iLast such that $|N(i)| = 2$. (Given MaxNum = 2, the condition $|N(i)| = 2$ is equivalent to Last($i$) > First($i$). All other sets, for which $|N(i)| = 1$, have Last($i$) = First($i$).) If Num2 is smaller than a chosen threshold value, such as Threshold = $n/16$, we may consider that it is not worthwhile to split the sets of the partition an additional time.

On the other hand, if Num2 > Threshold, then a final partition can be generated. It is relevant to observe how the algorithm handles the case where $|N(i)| = 1$. When the set $N(i)$ contains a single element, e.g., $N(i) = \{j\}$, then the rule for dividing $N(i)$ into $N_L(i)$ and $N_R(i)$ yields $N_L(i) = \{j\}$ and $N_R(i) = \varnothing$ if $i$ is odd, and $N_L(i) = \varnothing$ and $N_R(i) = \{j\}$ if $i$ is even. In the former case, only the assignments (2.1) and (2.3) are relevant, while in the latter case, only (2.2) and (2.4) are relevant.

The final operation of updating the partition can be skipped, since the only purpose of the partitions is to identify the assignments (2.1) to (2.4), and no additional assignments remain to be made.

**2.2 Implementation**

The algorithm can be implemented conveniently by observing there is no need to store the sets $N(i)$ at each step. Instead it suffices to record just two numbers, First($i$) and Last($i$), which identify $N(i)$ as given by $N(i) = \{j: \text{First}(i) \leq j \leq \text{Last}(i)\}$. The precise number of elements in a set $N(i)$ currently considered, which we call SetSize, is then given by SetSize = Last($i$) + 1 – First($i$).

$N(i)$ can thus be split by defining Split = $\lceil \text{SetSize}/2 \rceil$ if $i$ is odd and Split = $\lfloor \text{SetSize}/2 \rfloor$ if $i$ is even. This results in creating corresponding "First" and "Last" values for the sets $N_L(i)$ and $N_R(i)$ given as follows:

> SplitPoint = First($i$) + Split – 1
> First$_L(i)$ = First($i$)
> Last$_L(i)$ = SplitPoint
> First$_R(i)$ = SplitPoint + 1
> Last$_R(i)$ = Last($i$)

In the special case where $|N(i)| = 1$, the condition $N_L(i) = \varnothing$ or $N_R(i) = \varnothing$ results in First$_L(i)$ = Last$_L(i)$ + 1 or First$_R(i)$ = Last$_R(i)$ + 1, respectively. If the computer environment for implementing the method does not automatically bypass executing a loop of the form "For j = First to Last" under the condition First > Last, then this special situation needs to be handled separately.

One further type of streamlining is useful. The steps of the algorithm can be organized so that $N_L(i)$ and $N_R(i)$ need not be generated separately and then used to produce a new partition, but can instead be generated directly as new sets N(i) themselves. This requires the use of a vector Location(i) that identifies the location where the current "true" set N(i) is stored. More precisely, for i = 1 to iLast, the "First" and "Last" indexes that define N(i) are given by First(Loc) and Last(Loc) for Loc = Location(i).

The detailed form of the method is as follows, where we continue to make reference to vectors x(r) for r = 1 to rLast that may be used to store the successive vectors x' and x" generated. The only input for the method is the value Threshold that determines whether a last assignment should be made when the number Num2 of sets with |N(i)| = 2 is small (i.e., when Num2 ≤ Threshold).

**Max/Min Generation Method**
iLast = 1
First(1) = 1
Last(1) = n
% The next assignment remains invariant throughout the algorithm.
Location(1) = 1
% Generate the first two vectors x' and x" corresponding to x(0) and x(1).
x' = x
x" = Comp(x')
rLast = 1
% The iteration counter, Iter, is given a redundant bound of MaxIter = 100, noting that
    the method will handle a problem as large as $n = 2^k$ for k = MaxIter – 1.
MaxIter = 100
For Iter = 1 to MaxIter
    % Each iteration creates a new partition of N and associated vectors x' and x".
    % Update the vector index rLast for recording x(rLast) = x' and x(rLast+1) = x".
    rLast = rLast + 1
    For i = 1 to iLast
        % Split each set N(i) of the current partition for i = 1 to iLast.
        Loc = Location(i)
        SetSize = Last(Loc) + 1 – First(Loc)
        If i is odd then
            Split = $\lceil$SetSize/2$\rceil$
        Else
            Split = $\lfloor$SetSize/2$\rfloor$
        Endif
        SplitPoint = First(Loc) + Split – 1
        FirstL= First(Loc)
        LastL= SplitPoint
        FirstR = SplitPoint + 1
        LastR = Last(Loc)

% The next two loops carry out the assignments (1) – (4).
% (If FirstL > LastL or FirstR > LastR, the corresponding
   loop should be skipped.)
For j = FirstL to LastL
    $x_j' = 1 - x_j$
    $x_j'' = x_j$
Endfor (j)
For j = FirstR to LastR
    $x_j' = x_j$
    $x_j'' = 1 - x_j$
Endfor (j)
% First(Loc) = FirstL already is true
Last(Loc) = LastL
First(Loc + iLast) = FirstR
Last(Loc + iLast) = LastR
Endfor (i)
rLast = rLast + 1
If rLast ≥ rLim then *Stop*.
% Identify MaxNum = |N(1)|  (Location(1) = 1 is invariant).
MaxNum = Last(1) + 1 – First(1)
If MaxNum = 1 then
    % All vectors x' and x" have been generated. No need to update the final partition.
    Stop
Endif
% Update the partitions by updating the Location(i) array, to assure that
   Loc = Location(i) identifies where N(i) is stored for i = 1 to iLast.
For i = iLast to 1 (-1)   % i = iLast, iLast – 1, …, 1
    Loc = Location(i)
    Location(2i – 1) = Loc
    Location(2i) = Loc + iLast
Endfor (i)
iLast = 2iLast
If MaxNum = 2 then
    % Identify the number Num2 of sets having |N(i)| = 2. Don't need to use
      Loc = Location(i) since the order of the sets doesn't matter.
    Num2 = 0
    For i = 1 to iLast
        If Last(i) > First(i) then
            Num2 = Num2 + 1
        Endif
    Endfor (i)
    If Num2 ≤ Threshold then
        % Skip generating a final assignment. All relevant x' and x"
          vectors have been generated.
        Stop
    Endif

        Endif
Endfor (Iter)

The number of iterations of the method within the "For Iter = 1 to MaxIter" loop will equal $\lfloor \log_2 n \rfloor$ or $1 + \lfloor \log_2 n \rfloor$, depending on whether the method stops because Num2 ≤ Threshold. (Hence the algorithm produces either $2\lfloor \log_2 n \rfloor$ or $2 + 2\lfloor \log_2 n \rfloor$ vectors in total.)

**2.3 Illustration**

We illustrate the method applied to the case for N = {1, 2, …, 11}. The outcomes for each iteration are shown in a block headed by "Iter = 1," "Iter = 2," and so forth. Each set N(i) for the current Iter is identified within "{ }" brackets, immediately below the value shown for the associated index i. Following this are the symbols "L" and "R" identifying the sets $N_L(i)$ and $N_R(i)$, which are depicted in the form $\{(N_L(i))\ (N_R(i))\}$. Thus, for example, in the block for Iter = 3, the grouping {(7 8) (9 10 11)} beneath i = 2 discloses that $N_L(2)$ = {7, 8} and $N_R(2)$ = {9, 10, 11}.

Following the rules of the algorithm, when a set N(i) cannot be divided into two equal left and right halves, $N_L(i)$ is the "larger half" or "smaller half" according to whether i is odd or even. Consequently, for Iter = 3 and i = 2, where i is even, the set $N_L(2)$ is the smaller half of N(2) (containing 2 elements compared to the 3 elements of $N_R(2)$).

The vectors x' and x" illustrated are based on assuming the seed vector x is the 0 vector. Hence the first two vectors generated (not shown) are x' = (0, 0, …, 0) and x" = (1, 1, …, 1).

It should be pointed out that the partition shown at the beginning of each iteration is actually the one that is created by the updating operation at the conclusion of the preceding iteration. (The partition for Iter = 1 is the full set N, which is created as the initial N(1) outside the main loop, before Iter is assigned a value.) Listing the partitions in this way gives a better picture of the way the method operates, but provides a slight distortion concerning the termination condition. In particular, the value of MaxNum = |N(1)| shown at the beginning of each iteration is the MaxNum value identified by the algorithm at the conclusion of the preceding iteration. Consequently, as indicated below, the method terminates for this example at the end of Iter = 4, since the value MaxNum = 1 that triggers this termination is identified at the conclusion of this iteration.

```
Iter = 1
MaxNum = |N(1)| = 11
i =                    1
       {1 2 3 4 5 6 7 8 9 10 11}
               L             R
       {(1 2 3 4 5 6) (7 8 9 10 11)}
x' =   1 1 1 1 1 1   0 0 0 0 0
x" =   0 0 0 0 0 0   1 1 1 1 1

Iter = 2
MaxNum = |N(1)| = 6
i =              1              2
```

```
      {1 2 3 4 5 6}   {7 8 9 10 11}
         L     R         L     R
      {(1 2 3) (4 5 6)}  {(7 8) (9 10 11)}
x'  = 1 1 1  0 0 0    1 1  0 0 0
x'' = 0 0 0  1 1 1    0 0  1 1 1
```

Iter = 3  
MaxNum = |N(1)| = 3

```
i =      1          2         3        4
      {1 2 3}    {4 5 6}    {7 8}   {9 10 11}
       L  R      L  R      L  R     L    R
      {(1 2) (3)} {(4) (5 6)} {(7) (8)} {(9) (10 11)}
x'  = 1 1 0    1 0 0    1 0    1 0 0
x'' = 0 0 1    0 1 1    0 1    0 1 1
```

Iter = 4  
MaxNum = |N(1)| = 2

```
i =    1     2    3    4    5    6    7     8
     {1 2}  {3}  {4} {5 6} {7}  {8}  {9}  {10 11}
      L R    R    L   L R   L    R    L    L  R
     {(1) (2)} {(3)} {(4)} {(5) (6)} {(7)} {(8)} {(9)} {(10) (11)}
x'  = 1  0    0    0    1 0    1    0    1    1  0
x'' = 0  1    1    1    0 1    0    1    0    0  1
```

The method Stops at this point (by identifying MaxNum = |N(1)| = 1).

Appendix 2 gives a "balanced" variant of the Max/Min approach that more nearly assures the number of complemented and un-complemented elements are equal.

### 2.4 Modifying x' to Produce Different Numbers of Complemented Variables

We may modify the vector x' produced at each stage of the method by changing the treatment of every second or every third element such that $x_j' = 1 - x_j$ by instead setting $x_j' = x_j$ (e.g., setting $x_4'$, $x_7'$ and $x_{10}'$ equal to 0 in the last iteration of the illustration above if every second complemented element is changed, and setting $x_5'$ and $x_{10}'$ equal to 0 if every third complemented element is changed.). Similarly, we may replace every second or third element such that $x_j' = x_j$ by instead setting $x_j' = 1 - x_j$ (which sets $x_3'$ and $x_8'$ equal to 1 in the last iteration of the preceding illustration if every second such element is changed, and sets just $x_6'$ equal to 1 if every third such element is changed).

This departs from the Max/Min approach, which generates vectors consisting of approximately equal numbers of complemented and un-complemented elements, to produce vectors containing approximately 1/4 complemented and 3/4 un-complemented elements (or vice versa) if every second element designated element is changed, and approximately 1/3 complemented and 2/3 un-complemented elements (or vice versa) if every third designated element is changed. This additional collection of vectors, when added to the collection generated directly by the Max/Min approach, produces greater variety in the types of vectors produced, though with the outcome that the members of this larger collection are less diverse relative to each other.

The next section provides a "Augmented-Max/Min" approach that generates additional vectors by an easily implemented alternative rule.

### 3. Augmented-Max/Min Diversification Generator

The Augmented-Max/Min generation method, as in the case of the Max/Min generation approach, undertakes to subdivide N successively into k different approximately equal sized subsets, as k ranges over the k = 2, 4, 8, 16, …, where each subset is constructed to differ "as much as possible" from all others. Also, as in the Max/Min method, each subset contains approximately $\lfloor (n/k) + .5 \rfloor$ elements. Beyond this, however, the Augmented-Max/Min approach includes numbers of subsets halfway between these values, adding the values of k given by k = 3, 6, 12, … (hence k = $2^p$, $2^{p-1}$ + $2^p$, for p = 1, 2, 3, …). Each vector generated is accompanied by generating its complement, likewise as in the case of the Max/Min method.

For simplicity, as we have done in the illustration for the Max/Min method, our rules to describe the Augmented-Max/Min method will be framed as generating binary vectors from the seed vector x(0) = (0, 0, …, 0). Each vector $x_o'$ thus generated can be used to create a corresponding vector x' "derived from" an arbitrary seed vector x by setting $x_j'$ = $x_j$ if $x_{oj}'$ = 0, and $x_j'$ = 1 − $x_j$ if $x_{oj}'$ = 1. (In other words, x' results by complementing those components of x for which $x_{oj}'$ = 1, and leaving all remaining components of x unchanged.)

We denote the vectors generated by x((s)), for values of s = $\lfloor (n/k) + .5 \rfloor$ as k ranges over the values k = 2, 3, 4, 6, 8, 12, 16, …. The vector x((s)) consists of alternating strings 1's and 0's, each of size s – i.e., starting with s 1's, followed by s 0's, then s 1's, and so on. The final string within x((s)) contains s' ≤ s components where s' is the number remaining to give the vector x((s)) a total of n components. (Hence, s' = n – $\lfloor n/s \rfloor$·s, if $\lfloor n/s \rfloor$ is not an integer.) If n is a power of 2, and if we used only the values k = 2, 4, 8, … (that are likewise powers of 2), then the Augmented-Max/Min method would generate exactly the same collection of vectors as the Max/Min method.

To complete the description of the Augmented-Max/Min method, we impose a lower limit sLim = $\lfloor n^{.5} + .5 \rfloor$ on the size of the string s, noting that the value s = $\lfloor (n/k) + .5 \rfloor$ diminishes in size as k grows. In particular, we interrupt the process of generating the vectors x((s)) upon reaching the smallest value of s such that s > sLim. At this point, we complete the process by generating the final vectors x((s)) for the values of s given by s = sLim – 1, sLim – 2, …, 1.

### 3.1 Illustration

For n = 51, we begin with the values s given by s = $\lfloor (n/k) + .5 \rfloor$ for k = 2, 3, 4, 6 (since sLim = $\lfloor (51^{.5} + .5 \rfloor$ = 7). This yields

x((25)) consisting of 26 1's followed by 25 0's.
x((16)) consisting of 16 1's, then 16 0's, then 16 1's, then 16 0's, then 3 1's.

x((12)) consisting of 12 1's, then 12 0's, …, then 3 1's.
x((8)) consisting of 8 1's then 8 0's, …, then 3 1's

The sequence is then completed by

x((6)) consisting of 6 1's, then 6 0's, …, then 3 1's
x((5)) consisting of 5 1's, then 5 0's, …, then one 1.
…
x((1)) consisting of alternating 1's and 0's.

### 3.2 Extension by a Shifting Procedure

We enlarge the set x((s)) by creating an additional vector x°((s)) for each value of s > 1 by inserting $\lfloor s/2 \rfloor$ 0's at the start of x((s)), and drop the last $\lfloor s/2 \rfloor$ components of x((s)). (Hence x°((s)) "shifts" x((s)) to the right by $\lfloor s/2 \rfloor$ components.) We do not bother to consider x°((1)) since by definition this vector would shift x((1)) by 0 components. (The alternative of shifting x((1)) by 1 component is of no interest, since it just produces the complement of x((1)).)

As in the case of the x((s)) vectors, we also generate the complement of each x°((s)) vector. The collection produced by the Augmented-Max/Min method contains somewhat more than twice the number of vectors produced by the Max/Min method, and the simplicity of its rules commends it for use as an alternative approach. As in the case of the Max/Min method, alternating 1's in the x((s)) vectors may be replaced by 0's, or alternating 0's may be replaced by 1's, to produce different balances in the numbers of components of these vectors that are complemented and un-complemented.

The next section gives the algorithm that can be used in accompaniment with the foregoing algorithms to generate additional diversified vectors.

### 4. Expanded Diversification by Permutation Mappings

We now introduce a procedure that operates by mapping a given collection of vectors into one or more new collections that differ from the original collection in a manner consistent with the concept of diversity previously employed. This procedure incorporates a method proposed in Glover (1997) and applied by Campos, Laguna and Marti (2005) for generating diverse permutations, which we modify and then extend to provide a set of additional mappings. Adapted to the present context, the method expands the collection of vectors x(r), r = 0 to rAdd by adding vectors x(r) for r ranging from r = rAdd + 1 to rLim (the chosen limit on the total number of such vectors produced).

We make reference to a gap value g and a starting value s which s ranges from 1 to g. We also refer to an iteration index k that runs from 0 to a maximum value kMax = $\lfloor (n - s)/g \rfloor$ (hence identifying kMax to be the largest k such that the index j = s + kg satisfies j ≤ n). (The gap g and the indexes s and k are also used in the Progressive Gap method of Appendix 1.)

In the present setting we recommend setting $g = \lfloor n/2 \rfloor - 1$, which is particularly compatible with applying a recursive version of the current Permutation Mapping Algorithm in conjunction with the Max/Min Algorithm.

**4.1 Structure of the Diverse Permutations.**

The permutations generated derive from operating on a given vector of numbers $(1, \ldots, n)$, which we take to be the indexes of the variables $x_j$ for $j = 1$ to $n$. Within this context, we construct a permutation $P_n(g)$ of $(1, \ldots, n)$ by reference to a series of "sub-permutations" $P_n(g: s)$, for $s = 1$ to $g$, whose components are given by

$$P_n(g: s) = (s + kg: k = 0 \text{ to } kMax)$$

or equivalently

$$P_n(g: s) = (s, s + g, s + 2g, \ldots, s + kMaxg).$$

The sub-permutations $P_n(g: s)$ can be placed end to end in any order to create $P_n(g)$. However, we favor using the reverse order, hence creating

$$P_n(g) = (P_n(g: s): \text{for } s = g, g - 1, \ldots, 1).$$

**4.1.1 Illustration**

Consider the permutation $P_n(g)$ for the case $n = 14$ and $g = 6$ ($= \lfloor n/2 \rfloor - 1$). The sub-permutations of $P_n(g)$ are then

$P_n(g: 1) = (1\ \ 7\ \ 13)$
$P_n(g: 2) = (2\ \ 8\ \ 14)$
$P_n(g: 3) = (3\ \ 9)$
$P_n(g: 4) = (4\ \ 10)$
$P_n(g: 5) = (5\ \ 11)$
$P_n(g: 6) = (6\ \ 12)$

Assembling these sub-permutations in reverse order yields

$P_n(g) = (6\ \ 12\ \ 5\ \ 11\ \ 4\ \ 10\ \ 3\ \ 9\ \ 2\ \ 8\ \ 14\ \ 1\ \ 7\ \ 13)$

**4.2 Employing $P_n(g)$ as a Permutation Mapping**

We treat $P_n(g)$ as a mapping $M = (m(1), m(2), \ldots, m(n))$ that generates a vector $y(r)$ from a given vector $x(r)$ by defining $y_j(r) = x_{m(j)}(r)$. This gives rise to a new collection of diverse vectors $y(r)$, $r = 1$ to rLast from the original collection $x(r)$, $r = 1, \ldots,$ rLast in the following manner.

**Permutation Mapping Algorithm**

For r = 1 to rLast
    For j = 1 to n
        i = m(j)
        $y_j(r) = x_i(r)$
    Endfor (j)
Endfor (r)

When the Permutation MappingAlgorithm is applied to enlarge a current collection of vectors x(r). r = 1 to rAdd, the vector $y_j(r)$ above is replaced by $x_j(r + rAdd)$ (hence $x_j(r + rAdd) = x_i(r)$) as r ranges from 1 to rLast, followed by re-setting rAdd = rAdd + rLast. The process can be stopped at point when the value r + rAdd reaches the desired limit rLim on the total number of diverse vectors accumulated.

We now identify a way to go farther than a single application of the preceding algorithm.

**4.3 Recursive Permutation Mapping**

The mapping M = (m(1), m(2)…, m(n)) can be applied to any permutation P = (p(1), …, p(n)) of the indexes j = 1 to n, and not only to the initial permutation $P_o$ = (1, 2, …, n). We specifically define the mapping M(P) = P' = (p'(1), …, p'(n)) by

$$p'(j) = p(m(j)) \text{ for } j = 1, ..., m. \qquad (4.1)$$

The foregoing mapping therefore replaces the $j^{th}$ element of P' by the $m(j)^{th}$ element of P. Note that if P = $P_o$ = (1, 2, …, n) then P' = M(P) = M.

Since M itself can be any permutation, it follows that $P_o$ is the identity element with respect to all such mappings; i.e., $P_o(M) = M(P_o) = M$, taking M to be an arbitrary permutation. The inverse $M^{-1}$ of M, which yields $M^{-1}(M) = M(M^{-1}) = P_o$, and whose components are denoted by by writing $M^{-1}$ = $(m^{-1}(1), …, m^{-1}(n))$, can be identified from the following relationship:

$$m^{-1}(i) = j \text{ for } i = m(j), j = 1, …, n \qquad (4.2)$$

(hence $m^{-1}(m(j)) = j$ for all j, and noting that (4.2) also holds for i = 1, …,n, we also have $m(m^{-1}(i)) = i$ for all i).

In the present setting, we are only interested in permutations M of the form given by M = $P_n(g)$, for $P_n(g)$ as previously identified. We may illustrate the inverse mapping by reference to the illustration of section 4.1.1, where

    j  =  1   2   3   4   5   6   7   8   9  10  11  12  13  14
$P_n(g)$ = (6  12  5  11  4  10  3  9  2   8  14   1   7  13)

Then applying (4.2) for $M = P_n(g)$ to obtain the inverse, we have

$$j = \phantom{0}1\phantom{0}\phantom{0}2\phantom{0}\phantom{0}3\phantom{0}\phantom{0}4\phantom{0}\phantom{0}5\phantom{0}\phantom{0}6\phantom{0}\phantom{0}7\phantom{0}\phantom{0}8\phantom{0}\phantom{0}9\phantom{0}10\phantom{0}11\phantom{0}12\phantom{0}13\phantom{0}14$$
$$M^{-1} = (12\phantom{0}\phantom{0}9\phantom{0}\phantom{0}7\phantom{0}\phantom{0}5\phantom{0}\phantom{0}3\phantom{0}\phantom{0}1\phantom{0}13\phantom{0}10\phantom{0}\phantom{0}8\phantom{0}\phantom{0}6\phantom{0}\phantom{0}4\phantom{0}\phantom{0}2\phantom{0}14\phantom{0}11)$$

($M^{-1}$ may be constructed conveniently using visual cues by looking for the successive indexes $i = 1, \ldots, n$ such that $m(j) = i$.)

### 4.3.1 Recursive Use of M

To use M recursively, we start by applying M to $P_o$ obtain $M(P_o) = M$ as the first permutation of a series. This first M, which we denote by $M^1$, is the one used to generate $y_j(r) = x_i(r)$, for $i = m(j)$, $j = 1$ to n, by the Permutation Mapping Algorithm. Then we apply the mapping M again to obtain the mapping $M(M(P_o))$, or $M^2(P_o) = M^2$, where we define $M^2 = M(M)$. Now apply the Permutation Mapping Algorithm with M replaced by $M^2$ in its description. (I.e., we replace $m(j)$ in this algorithm by $m^2(j)$, where $M^2 = (m^2(1), \ldots, m^2(n))$.) This is equivalent to redefining $x(r)$ to be the vector $y(r)$ produced by the first application of the Permutation Mapping Algorithm, followed by applying the algorithm in its original form (without replacing M by $M^2$) to the resulting new $x(r)$ vector.

In a similar manner, we may generate the mapping $M^3 = M(M(M)) = M(M^2)$ and apply the Permutation Mapping Algorithm with M replaced by $M^3 = (m^3(1), \ldots, m^3(n))$. Again, equivalently, this corresponds to applying the Permutation Mapping Algorithm unchanged to the "updated" vector $x(r)$ (which is the new vector $y(r)$ obtained from the preceding pass). The recursive use of M in this fashion is motivated by the expectation that each step should create a useful diversification relative to the vector last produced, given that M is designed to create such diversification relative to the permutation $P_o$ which is an arbitrary initial indexing for the variables.

Eventually, for some value $h \geq 1$ we obtain a "next" mapping $M^{h+1} = M(M^h)$ that yields the initial vector $P_o = (1, \ldots, n)$ as its outcome, and the process cycles. The relationship $M(M^h) = P_o$ discloses that $M^h$ is in fact the inverse mapping $M^{-1}$. This further implies that we can obtain the same collection of $y(r)$ vectors by starting with $M^{-1} (= M^h)$, then continuing with $M^{-2} = M^{-1}(M^{-1})$ $(= M^{h-1})$, until finally reaching $M^{-h} (= M^1 = M)$. In other words, starting with $M^{-1}$ generates the same collection of $y(r)$ vectors as starting with M, but in reverse order. Consequently, $M^{-1}$ is on an equal footing with M as a diversifying permutation mapping. (The vector produced by reversing the order of the components of M does not have this same footing.)

When applying the mapping M recursively as indicated, the number of different $y(r)$ vectors that can be produced before reaching the "last" mapping $M^h$ grows rapidly with the value of n (using the definition of $M = P_n(g)$). Consequently, the limit rLim on the total number of vectors generated may be reached long before cycling occurs. (Other definitions of M can potentially produce larger numbers of vectors before cycling, but our primary goal remains that of producing a diverse collection rather than a collection containing numerous elements.)

### 4.4 Illustrated Use of Recursion

We illustrate this recursive process for n = 9, where only a relatively small number of mappings are generated before cycling. For greater scope, we apply the mapping $M = P_n(g)$ simultaneously to all of the vectors produced by the Max/Min Algorithm of Section 2. For n = 9 we have $g = \lfloor n/2 \rfloor - 1 = 3$, and hence

$P_9(1: 3) = (1 \ 4 \ 7)$
$P_9(2: 3) = (2 \ 5 \ 8)$
$P_9(3: 3) = (3 \ 6 \ 9)$

to yield

$M = P_n(g) = (3 \ 6 \ 9 \ 2 \ 5 \ 8 \ 1 \ 4 \ 7)$

The first (upper left) section of Table 1 below shows the 8 vectors produced by the Max/Min Algorithm, and lists the indexes j = 1 to n, the mapping M and the initial vector $P_o$ (shown as P0) above them. The next section, immediately below the first, shows the corresponding vectors upon applying M to the first section. Thus $P_o$ is replaced by $M^1$ (shown as M1), and the vectors listed as 9 through 16 are the result of applying M to the vectors listed as 1 through 8.

The third section likewise results by applying M to the second section, replacing $M^1$ by $M^2$ (shown as $M^2$) and producing the vectors 17 through 24 from the corresponding vectors 9 through 16. The next section, which applies M once more to yield $M^3$ (shown as M3), is the final pass of the recursive process, as may be verified by noting that $M^3$ is in fact the inverse $M^{-1}$ of M. The table shows the additional step that produces the vector $M^4 = P_o$, and causes all of the resulting vectors to be the same as in the first section of the table, though of course this step is not necessary.

|       | 1 | 2 | 3 | 4 | 5 | 6 | 7 | 8 | 9 | index |
|-------|---|---|---|---|---|---|---|---|---|-------|
|       | 3 | 6 | 9 | 2 | 5 | 8 | 1 | 4 | 7 | M     |
|       | 1 | 2 | 3 | 4 | 5 | 6 | 7 | 8 | 9 | P0    |
| 1     | 1 | 1 | 1 | 1 | 1 | 0 | 0 | 0 | 0 |       |
| 2     | 0 | 0 | 0 | 0 | 0 | 1 | 1 | 1 | 1 |       |
| 3     | 1 | 1 | 1 | 0 | 0 | 1 | 1 | 0 | 0 |       |
| 4     | 0 | 0 | 0 | 1 | 1 | 0 | 0 | 1 | 1 |       |
| 5     | 1 | 1 | 0 | 1 | 0 | 1 | 0 | 1 | 0 |       |
| 6     | 0 | 0 | 1 | 0 | 1 | 0 | 1 | 0 | 1 |       |
| 7     | 1 | 0 | 0 | 1 | 0 | 1 | 0 | 1 | 0 |       |
| 8     | 0 | 1 | 1 | 0 | 1 | 0 | 1 | 0 | 1 |       |

|       | 1 | 2 | 3 | 4 | 5 | 6 | 7 | 8 | 9 | index |
|-------|---|---|---|---|---|---|---|---|---|-------|
|       | 3 | 6 | 9 | 2 | 5 | 8 | 1 | 4 | 7 | M     |
|       | 3 | 6 | 9 | 2 | 5 | 8 | 1 | 4 | 7 | M1    |
| 9     | 1 | 0 | 0 | 1 | 1 | 0 | 1 | 1 | 0 |       |
| 10    | 0 | 1 | 1 | 0 | 0 | 1 | 0 | 0 | 1 |       |
| 11    | 1 | 1 | 0 | 1 | 0 | 0 | 1 | 0 | 1 |       |
| 12    | 0 | 0 | 1 | 0 | 1 | 1 | 0 | 1 | 0 |       |
| 13    | 0 | 1 | 0 | 1 | 0 | 1 | 1 | 1 | 0 |       |
| 14    | 1 | 0 | 1 | 0 | 1 | 0 | 0 | 0 | 1 |       |
| 15    | 0 | 1 | 0 | 0 | 0 | 1 | 1 | 1 | 0 |       |
| 16    | 1 | 0 | 1 | 1 | 1 | 0 | 0 | 0 | 1 |       |

|       | 1 | 2 | 3 | 4 | 5 | 6 | 7 | 8 | 9 | index |
|-------|---|---|---|---|---|---|---|---|---|-------|
|       | 3 | 6 | 9 | 2 | 5 | 8 | 1 | 4 | 7 | M     |
|       | 9 | 8 | 7 | 6 | 5 | 4 | 3 | 2 | 1 | M2    |
| 17    | 0 | 0 | 0 | 0 | 1 | 1 | 1 | 1 | 1 |       |
| 18    | 1 | 1 | 1 | 1 | 0 | 0 | 0 | 0 | 0 |       |
| 19    | 0 | 0 | 1 | 1 | 0 | 0 | 1 | 1 | 1 |       |
| 20    | 1 | 1 | 0 | 0 | 1 | 1 | 0 | 0 | 0 |       |
| 21    | 0 | 1 | 0 | 1 | 0 | 1 | 0 | 1 | 1 |       |
| 22    | 1 | 0 | 1 | 0 | 1 | 0 | 1 | 0 | 0 |       |
| 23    | 0 | 1 | 0 | 1 | 0 | 1 | 0 | 0 | 1 |       |
| 24    | 1 | 0 | 1 | 0 | 1 | 0 | 1 | 1 | 0 |       |

|       | 1 | 2 | 3 | 4 | 5 | 6 | 7 | 8 | 9 | index |
|-------|---|---|---|---|---|---|---|---|---|-------|
|       | 3 | 6 | 9 | 2 | 5 | 8 | 1 | 4 | 7 | M     |
|       | 7 | 4 | 1 | 8 | 5 | 2 | 9 | 6 | 3 | M3    |
| 25    | 0 | 1 | 1 | 0 | 1 | 1 | 0 | 0 | 1 |       |
| 26    | 1 | 0 | 0 | 1 | 0 | 0 | 1 | 1 | 0 |       |
| 27    | 1 | 0 | 1 | 0 | 0 | 1 | 0 | 1 | 1 |       |
| 28    | 0 | 1 | 0 | 1 | 1 | 0 | 1 | 0 | 0 |       |
| 29    | 0 | 1 | 1 | 1 | 0 | 1 | 0 | 1 | 0 |       |
| 30    | 1 | 0 | 0 | 0 | 1 | 0 | 1 | 0 | 1 |       |
| 31    | 0 | 1 | 1 | 1 | 0 | 0 | 0 | 1 | 0 |       |
| 32    | 1 | 0 | 0 | 0 | 1 | 1 | 1 | 0 | 1 |       |

|       | 1 | 2 | 3 | 4 | 5 | 6 | 7 | 8 | 9 | index |
|-------|---|---|---|---|---|---|---|---|---|-------|
|       | 3 | 6 | 9 | 2 | 5 | 8 | 1 | 4 | 7 | M     |
|       | 1 | 2 | 3 | 4 | 5 | 6 | 7 | 8 | 9 | M4    |
| 1     | 1 | 1 | 1 | 1 | 0 | 0 | 0 | 0 |   |       |
| 2     | 0 | 0 | 0 | 0 | 0 | 1 | 1 | 1 | 1 |       |
| 3     | 1 | 1 | 1 | 0 | 0 | 1 | 1 | 0 | 0 |       |
| 4     | 0 | 0 | 0 | 1 | 1 | 0 | 0 | 1 | 1 |       |
| 5     | 1 | 1 | 0 | 1 | 0 | 1 | 0 | 1 | 0 |       |
| 6     | 0 | 0 | 1 | 0 | 1 | 0 | 1 | 0 | 1 |       |
| 7     | 1 | 0 | 0 | 1 | 0 | 1 | 0 | 1 | 0 |       |
| 8     | 0 | 1 | 1 | 0 | 1 | 0 | 1 | 0 | 1 |       |

**Table 1: Simultaneous Mapping of All 8 Vectors Produced by the Max/Min Algorithm for n = 9.**

The next section examines additional ways to generate diverse vectors, which can also be processed by the recursive mapping process to produce larger numbers of vectors.

## 5. Diversified Vectors from Balanced Sub-Vectors

An auxiliary type of diversification approach results from a construction that is approximately the inverse of the one underlying the Max/Min Generation method. Instead of doing a "successive binary partitioning" of the index set for a seed vector, as a basis for identifying variables to complement, we start from the other end and employ a constructive process to achieve an objective similar to that pursued by the Max/Min Generation method.

**5.1 Sub-Vector Coverage**

Let $y = (y_1, \ldots, y_p)$ denote a p-dimensional sub-vector that we seek to incorporate within a vector x' by repeating y multiple times within x'. We will produce a collection Y of these p-dimensional sub-vectors, and use each $y \in Y$ to build a different vector x'. Evidently we want the vectors y in Y to differ from each other, since this will assure the resulting vectors x' will likewise differ, and if p is not large, then the differences between the vectors y in Y will be magnified in the vectors x' since the latter will differ over a larger number (and proportion) of their components. To facilitate the analysis, we again suppose the seed vector x is the 0 vector and understand that the assignment $x_j' = 0$ corresponds to setting $x_j' = x_j$ and the assignment $x_j' = 1$ corresponds to setting $x_j' = 1 - x_j$.

For the purpose of keeping p relatively small, we start by considering values of p in the range from 3 to 7. For a given value of p, we obtain a "maximum coverage" of the sub-space associated with the vectors $y = (y_1, \ldots, y_p)$ in Y if these vectors constitute all $2^p$ binary sub-vectors of dimension p (hence yielding $|Y| = 2^p$ with a cardinality ranging from 8 to 128 for the indicated small p values). This maximum sub-space coverage derives from the obvious fact that no other collection of p-dimensional sub-vectors succeeds to matching every 0-1 vector possibility in the sub-space. However, the vectors y in Y by themselves are not particularly attractive as components to be incorporated in the vectors x', because Y does not come close to satisfying the balanced diversity criterion which would require each of its members $y \in Y$ to have approximately half of its entries 1 and half 0. In fact, by satisfying the maximum sub-vector coverage property, Y conflicts with the balanced diversity criterion to the greatest extent possible.

To remedy this shortcoming, we treat each vector y in Y as the first half of a larger vector containing 2p components. Denoting a specific vector $y \in Y$ by y', we choose the second half of the 2p component vector to consist of the complement y" of y' (which is also in Y). Then the "double length" vector $y^2 = (y', y'')$ possesses the desired property of containing half 0's and half 1's and yet the collection of such $y^2$ vectors satisfies a relaxed form of the maximum sub-vector coverage property in that both of the halves y' and y" of $y^2$ satisfy this property in relation to p-dimensional vectors as y' (and hence y") ranges over the $2^p$ vectors in Y to produce $y^2$.

We replicate this new $y^2$ vector as many times as possible to generate a n-vector $x' = (y^2, y^2, y^2, \ldots, y^2)$, understanding that the final $y^2$ is truncated as necessary to permit x' to have n components. Then x' will also meet the balanced diversity criterion of containing roughly half 0's and half 1's (as will its complement x").

Performing this same doubling operation with each of the $2^p$ vectors y' in Y, we create $2^p$ corresponding vectors $y^2 = (y', y'')$, and thus produce in turn $2^p$ vectors of the form $x' = (y^2, y^2, y^2, \ldots, y^2)$.

The ability to choose p relatively small results from the fact that the $2^p$ vectors x' (and the associated $2^p$ vectors x") will constitute a sufficiently large number to provide as many of these vectors as desired while p retains a modest value. We can also choose different values of p, and generate different composite vectors $y^2 = (y', y'')$ to build up different x' vectors.

**Illustration**

This construction is illustrated for p = 3 by listing the $2^p = 8$ vectors y' ∈ Y on the left below, and matching each with its complement y" on the right.

```
 h     y'      y"
----  -------  --------
 1    1 1 1    0 0 0
 2    1 1 0    0 0 1
 3    1 0 1    0 1 0
 4    1 0 0    0 1 1
 5    0 1 1    1 0 0
 6    0 1 0    1 0 1
 7    0 0 1    1 1 0
 8    0 0 0    1 1 1
```

For purposes of generating these y' and y" vectors, note that the vectors y' in the left column above correspond to listing the binary numbers from 0 to 7 in a bottom-to-top sequence and the vectors y" in the right column correspond to listing these same numbers in a top-to-bottom sequence. Accordingly, a convenient way to generate such vectors is to refer to the binary numbers that correspond to the vectors y' and then, upon listing them in reverse order, to create the vectors y" that correspond to the binary numbers in this reverse ordering.

Finally, upon coupling these y' and y" vectors to yield the 8 vectors of the form $y^2 = (y', y'')$, we obtain the following 8 vectors $x' = (y^2, y^2, \ldots, y^2)$, where we insert the symbol "|" to depict the separation between successive $y^2$ vectors.

(1, 1, 1, 0, 0, 0, | 1, 1, 1, 0, 0, 0, | 1, 1, 1, 0, 0, 0, | …)
(1, 1, 0, 0, 0, 1, | 1, 1, 0, 0, 0, 1, | 1, 1, 0, 0, 0, 1, |…)
(1, 0, 1, 0, 1, 0, | 1, 0, 1, 0, 1, 0, | 1, 0, 1, 0, 1, 0, |…)
. . . . . .

(0, 0, 0, 1, 1, 1, | 0, 0, 0, 1, 1, 1, | 0, 0, 0, 1, 1, 1, |…)

Such a collection may either be used by itself or added to those generated by the other algorithms of this paper to provide additional vectors (noting that some of the vectors of the current collection can also duplicate some of those generated by the other algorithms).

**Generating Vectors with Different Balances Between 1's and 0's**

As in the case of the Max/Min collection, we can generate vectors consisting of a different ratio of 1's and 0's. In addition to modifying vectors already generated by assigning some of their components the opposite of the value previously assigned, we can also produce a different form of variation in the numbers of 1's and 0's in the following manner.

Each vector pair y', y" is extended to become a triple y', y", y°, where y° is defined by setting $y_j^o = y_j'$ for $j \leq \lfloor p/2 \rfloor$ and $y_j^o = y_j"$ for $j > \lfloor p/2 \rfloor$. (Equivalently, y° complements the "second half" of the y' vector, leaving the first half unchanged.) Hence y° results by complementing roughly half the components of each of y' and y", and thus is "maximally different" from these two vectors. (This effect is best achieved when p is chosen to be an even number.) We make use of this string of 3p elements by assembling each of its $2^p$ instances end to end to produce $2^p$ different x' vectors.

The number of 1's and 0's will vary by adding from 0 to p additional 1's to each vector. (Most vectors will add p/2 new 1's, then the next largest number of vectors will add p/2 +1 or p/2 – 1 new 1's, etc.. For example, when p = 4, producing $2^p = 16$ different vectors, the number of vectors that add k 1's will be 1 for k = 0, 4 for k = 1, 6 for k = 2, 4 for k = 3 and 1 for k = 4.)

## 6. Conclusions

Strategies that generate meaningful collections of diverse vectors are highly desirable in metaheuristic optimization. As a foundation for creating such collections, we have shown how various forms of a Max/Min principle lead to diversification methods that can be usefully refined and generalized by augmentation and shifting procedures, and by special types of permutation mappings. Working backward, we also show how to achieve useful forms of diversification by a simple constructive approach to generate balanced sub-vectors.

Our methods motivate future research to apply them in the presence of constraints that are imposed to achieve intensification as well as diversification goals, as by bounding admissible objective function values or by setting limits on admissible distances from previous high quality solutions, and using supporting methods such as strategic oscillation that alternately drive the search to violate such limits and then to enforce them again by manipulating neighborhoods and search directions.

An instance of this type of extension consists of methods for generating diverse vectors that yield a selected number of elements in particular subsets equal to 1, using the Max/Min approach as an internal routine. Such methods can be useful in metaheuristic intensification strategies where it can be valuable to look for new solutions in which specified subsets of variables have approximately the same number of elements equal to 1 as in the best solutions. Joining such an approach with clustering strategies, and identifying different subsets of variables that may be relevant in different clusters, provides an area for further refinement.

E. Mayoraz and M. Moreira (1999) "Combinatorial Approach for Data Binarization," chapter in *Principles of Data Mining and Knowledge Discovery,* Volume 1704 of the series *Lecture Notes in Computer Science*, pp 442-447.

**Appendix 1: The Progressive Gap (PG) Method**

We slightly modify the original description of the Progressive Gap method to clarify its main components and to give a foundation for the Extended PG method described below.

**Notation for the PG Method**

$g$ = a gap value
$s$ = a starting index
$k$ = an increment index

**Method Overview**

Starting with the seed vector x, successive vectors x' are generated by complementing specific components $x_j$ of x. A gap value g is used that iteratively varies over the range g = 1 to gMax, where gMax = $\lfloor n^{.5} + .5 \rfloor$.[1] Then, for each gap g, a starting index s iterates from s = 1 to sLim, where sLim = g except in the special case where g = 2 where sLim is restricted to 1 (to avoid a duplication among the vectors x' generated).

From the initial assignment x' = x, the method sets $x_j'$ = 1 − $x_j$ for the index j = s + kg, as the increment index k ranges from 0 to kMax = $\lfloor (n - s)/g \rfloor$. Thus, $x_j'$ receives this complemented value of $x_j$ for j = s, s + g, s + 3g, …, thereby causes each j to be separated from the previous j by the gap of g. (The actual gap between two successive values of j is thus g − 1. For example, when g = 1, the values j and j + g = j + 1 are adjacent, and in this sense have a "0 gap" between them.) The indicated formula for the maximum value of k sets kMax as large as possible, subject to assuring j does not exceed n (when j attains its largest value j = s + kMax·g). Each time a vector x' is generated, the corresponding vector x" = Comp(x') is also generated. This simple pattern is repeated until no more gaps g or starting values s remaining to be considered.

**PG Algorithm**

rLast = 0
gMax = $\lfloor n^{.5} + .5 \rfloor$
% Iterate over gap values g.
For g = 1 to gMax
       % Choose the max starting index sLim to be the same as the gap g unless g = 2.

---

[1] A different limiting value for g is proposed in Glover (1997), consisting of gMax = $\lfloor n/5 \rfloor$. The rationale for this upper limit in both cases is based on the fact that as g grows, the difference between x and x' becomes smaller, and hence a bound is sought that will prevent x' from becoming too similar to x.

If g = 2 then
                     sLim = 1
              else
                     sLim = g
              Endif
              % Iterate over starting values s
              For s = 1 to sLim
                     % Identify the largest value kMax for the increment index k so that the index j of $x_j$
                         given by j = s + kg will not exceed n.
                     kMax = $\lfloor (n - s)/g \rfloor$
                     % Increment the index rLast for identifying the vectors currently generated
                         as x(r) for r = 0 to rLast.
                     rLast = rLast + 1
                     % Begin generating the new vector x(rLast) = x'.
                     x' = x
                     % Start j at the starting index s.
                     j = s
                     For k = 0 to kMax
                            % For each value k, implicitly j = s + kg
                            $x_j$' = 1 − $x_j$
                            % Insert a gap of g between the current j and the next j
                            j = j + g
                     Endfor (k)
                     % the new vector x(rLast) = x' is now completed. Increment rLast and generate
                         the complement x" of x' to implicitly identify x(rLast) = x" for the next
                         value of rLast.
                     rLast = rLast + 1
                     x" = Comp(x')
                     If rLast ≥ rLim then *Stop*
              Endfor (s)
Endfor (g)

*Remark*: The method can avoid generating x" = Comp(x') when x' is the first vector generated (i.e., x' = x(1)), since in this case Comp(x') = x, thus yielding the seed vector (x(0)).

To illustrate for the case where the seed vector is x = (0, 0, …, 0), the procedure generates the following vectors x' for the sampling of values shown for the starting index s and the gap g. Note that the vector x' for s = 2, g = 2 (marked with a "*" below) duplicates the complement of the x' vector for s = 1, g = 2. This is the reason the algorithm restricts the value sLim to 1 when g = 2, thus causing the vector for s = 2, g = 2 to be skipped.

s = 1, g = 2:  (1 0 1 0 1 0 1 0 1 0 …)
s = 1, g = 3:  (1 0 0 1 0 0 1 0 0 1 …)
s = 2, g = 2:  (0 1 0 1 0 1 0 1 0 0 …) *
s = 2, g = 3:  (0 1 0 0 1 0 0 1 0 0 …)
s = 3, g = 2:  (0 0 1 0 1 0 1 0 1 0 …)

s = 3, g = 3: (0 0 1 0 0 1 0 0 1 0 …)

**Extended Version**

The Extended PG Method can be used to generate a larger number of points, and also provides an additional form of variation in the vectors generated.

The extended version of the Progressive Gap Method is for situations where the basic version of the method provides fewer points than desired.

**Brief Overview.**

The extended method "fills in spaces" between successive j values that determine the assignment $x_j' = 1 - x_j$. The method makes this assignment for a string of j values from $j = j_1$ to $j_2$, where $j_2$ is chosen to leave an unassigned position between $j_2$ and the next value of $j_1$ given by $j_1 = j_j + g$. Consequently, $j_2 = j_1 + g - 2$ (and the method chooses $j_2 = j_1$ until g > 2.)

The resulting algorithm avoids referring to a starting index s to identify the location of the "first j value" at which $x_j' = 1 - x_j$. Instead, the starting value is always j = 1. This results from the fact that the complements x" produced for the x' vectors automatically include all of the vectors x' that would be derived by using different starting indexes s.

The extended algorithm is stated as follows.

**Extended PGAlgorithm**

rLast = 0
gMax = $\lfloor n^{.5} + .5 \rfloor$
% Iterate over gap values g.
For g = 1 to gMax
    kMax = $\lfloor (n - 1)/g \rfloor$
    rLast = rLast + 1
    % Begin generating the new vector x(rLast) = x'.
    x' = x
    % Start $j_1$ at the value 1.
    $j_1$ = 1
    % identify the max value $\Delta g_{max}$ that is added to $j_1$ to produce $j_2$
    If g = 1 then
        $\Delta g_{max}$ = 0
    Else
        $\Delta g_{max}$ = g − 2
    Endif
    For $\Delta g$ = 0 to $\Delta g_{max}$
        For k = 0 to kMax
            % For each value k, implicitly $j_j$ = 1 + kg
            $j_2 = j_1 + \Delta g$

```
                For j = j₁ to j₂
                        xⱼ' = 1 − xⱼ
                Endfor (j)
                % Insert a gap of g between the current j₁ and the next j₁
                j₁ = j₁ + g
        Endfor (k)
        % the new vector x(rLast) = x' is now completed. Increment rLast and generate
            the complement x" of x' (implicitly identifying x(rLast) = x" for the next
            value of rLast.
        rLast = rLast + 1
        x" = Comp(x')
        If rLast ≥ rLim then Stop
    Endfor (Δg)
Endfor (g)
```

The PG Algorithm can be extended in additional ways, but we restrict attention to the preceding approach as the primary variation. Combining either the PG Algorithm its extension with Algorithm 3 will succeed in producing an additional collection of diversified vectors if still more such vectors are sought.

**Appendix 2: A "Balanced" Variant of the Max/Min Algorithm**

The idea underlying the Balanced Variant of Algorithm 2 is to assure that sets $N(i)$ with an odd number of elements SetSize are split so that $\lfloor SetSize/2 \rfloor$ of their elements go into $N_L(i)$ when an odd number of such sets have been encountered and $\lceil SetSize/2 \rceil$ of their elements go into $N_L(i)$ when an even number of such sets have been encountered. The rule is applied anew at each iteration (each successive value of Iter), when creating a new partition from the current sets $N(i)$ for $i = 1$ to iLast.

The "balanced" terminology comes from the fact that this approach will tend to balance the number of variables $x_j$ that are complemented and not complemented to produce the vector x' generated on the current iteration. When this approach is not used, the order in which the current $N(i)$ sets occur could cause each set with $|N(i)|$ odd to be split in the same way, putting $\lceil SetSize/2 \rceil$ (or $\lfloor SetSize/2 \rfloor$) elements in $N_L(i)$, thus causing the number of complemented $x_j$ to exceed the number of $x_j$ not complemented (or vice versa).

When the Balanced Variant is used, the final assignment to be made (following the determination that MaxNum = 2) has a simple form that allows x' and x" to be created by the following shortcut step.

$$x_j' = 1 - x_j \quad \text{if } j \text{ is odd} \tag{1'}$$
$$x_j' = x_j \quad \text{if } j \text{ is even} \tag{2'}$$

and

$$x_j" = x_j \quad \text{if } j \text{ is odd} \tag{3'}$$
$$x_j" = 1 - x_j \quad \text{if } j \text{ is even} \tag{4'}$$

Consequently, when MaxNum = 2, the method immediately makes this simplified final assignment and then stops.

The detailed form of this approach is as follows. A logical variable named OddSet keeps track of whether an even odd number of sets with |N(i)| odd have been encountered.

**Balanced Variant of the Max/Min Generation Method**
iLast = 1
First(1) = 1
Last(1) = n
Location(1) = 1
% Generate the first two vectors x' and x" corresponding to x(0) and x(1).
x' = x
x" = Comp(x')
rLast = 1
MaxIter = 100
For Iter = 1 to MaxIter
    % Each iteration creates a new partition of N and associated vectors x' and x".
    % Update the vector index rLast for recording x(rLast) = x' and x(rLast+1) = x".
    rLast = rLast + 1
    % Initialize the logical variable OddSet to keep track of whether an even or odd number of
        sets N(i) have been encountered with SetSize = |N(i)| odd.
    OddSet = True
    For i = 1 to iLast
        % Split each set N(i) of the current partition for i = 1 to iLast.
        Loc = Location(i)
        SetSize = Last(Loc) + 1 − First(Loc)
        If SetSize is odd then
            If OddSet = True then
                Split = $\lfloor$SetSize/2$\rfloor$
                OddSet = False
            Else
                Split = $\lceil$SetSize/2$\rceil$
                OddSet = True
            Endif
        Else
            Split = SetSize/2
        Endif
        SplitPoint = First(Loc) + Split − 1
        FirstL= First(Loc)
        LastL= SplitPoint
        FirstR = SplitPoint + 1
        LastR = Last(Loc)
        % The next two loops carry out the assignments (1) – (4).
        % (If FirstL > LastL or FirstR > LastR, the corresponding

loop should be skipped.)
For j = FirstL to LastL
    $x_j' = 1 - x_j$
    $x_j'' = x_j$
Endfor (j)
For j = FirstR to LastR
    $x_j' = x_j$
    $x_j'' = 1 - x_j$
Endfor (j)
% First(Loc) = FirstL already is true
Last(Loc) = LastL
First(Loc + iLast) = FirstR
Last(Loc + iLast) = LastR
Endfor (i)
rLast = rLast + 1
If rLast ≥ rLim then *Stop*.
% Identify MaxNum = |N(1)|  (Location(1) = 1 is invariant).
MaxNum = Last(1) + 1 − First(1)
If MaxNum = 1 then
    % All vectors x' and x" have been generated. No need to update the final partition.
    Stop
Endif
% Update the partitions by updating the Location(i) array, to assure that
% Loc = Location(i) identifies where N(i) is stored for i = 1 to iLast.
For i = iLast to 1 (-1)   % i = iLast, iLast − 1, …, 1
    Loc = Location(i)
    Location(2i − 1) = Loc
    Location(2i) = Loc + iLast
Endfor (i)
iLast = 2iLast
If MaxNum = 2 then
    % Identify the number Num2 of sets having |N(i)| = 2. Don't need to use
       Loc = Location(i) since the order of the sets doesn't matter.
    Num2 = 0
    For i = 1 to iLast
        If Last(i) > First(i) then
            Num2 = Num2 + 1
        Endif
    Endfor (i)
    If Num2 ≤ Threshold then
        % Skip generating a final assignment. All relevant x' and x"
           vectors have been generated.
        Stop
    Else
        % Generate the shortcut assignment (1') to (4').
        For j = 1 to n

If j is odd then
$$x_j' = 1 - x_j$$
$$x_j'' = x_j$$
Else
$$x_j' = x_j$$
$$x_j'' = 1 - x_j$$
Endif
Endfor
Stop
Endif
Endif
Endfor (Iter)

**Appendix 3: Strongly Balanced Vector Generation**

We consider a recursive process to generate diverse vectors that are not only composed of approximately half 0's and half 1's, but that additionally are *strongly balanced* in the sense that every successive pair of elements consists of a single 0 and a single 1. We start with the case for p = 2 and consider just the 2 vectors that contain exactly one 0 and one 1, which are complements of each other:

y(1) = (1,0) and y(2) = (0,1).

We could use these vectors by themselves to generate the two x vectors given by x(1) = (y(1), y(1), …) and x(2) = (y(2), y(2), …), which also are complements of each other.

Now we consider all ways of pairing these two vectors, thus obtaining all vectors of the form (y(p), y(q)) for p, q = 1, 2 (i.e., (y(1), y(1)), (y(1), y(2)), …, etc.) From this we obtain the 4 new vectors

y(1) = (1, 0, 1, 0), y(2) = (1, 0, 0, 1), y(3) = (0, 1, 1, 0), y(4) = (0, 1, 0, 1)

The complement of each of these vectors is also contained in the collection generated. (For example, y(1) and y(4) are complements, and y(2) and y(3) are complements.) Moreover, these y vectors satisfy the strongly balanced property where every successive two components of these vectors consists of one 0 and one 1.

Again, we can form the vectors x(h) = (y(h), y(h), ….) for h = 1 to 4 and the complement of each vector is likewise in the collection. (This holds even if the last y(h) vector in each x(h) must be truncated so that x(h) has dimension n.) Similarly, every two successive components of each vector consists of one 0 and one 1, although if n is odd there will not be a final "second component" to pair with $x_n(h)$.

To take this process one step farther, we combine the vectors y(1) through y(4) to produce all possible pairs (y(p), y(q)) for p, q = 1, 2, 3, 4. The 4 x 4 = 16 resulting combinations are shown below.

```
 h      y(p)        y(q)
----  ----------  ----------
  1   1 0 1 0     1 0 1 0
  2   1 0 1 0     0 1 0 1
  3   1 0 1 0     0 1 1 0
  4   1 0 1 0     1 0 0 1
  5   0 1 0 1     1 0 1 0
  6   0 1 0 1     0 1 0 1
  7   0 1 0 1     0 1 1 0
  8   0 1 0 1     1 0 0 1
  9   0 1 1 0     1 0 1 0
 10   0 1 1 0     0 1 0 1
 11   0 1 1 0     0 1 1 0
 12   0 1 1 0     1 0 0 1
 13   1 0 0 1     1 0 1 0
 14   1 0 0 1     0 1 0 1
 15   1 0 0 1     0 1 1 0
 16   1 0 0 1     1 0 0 1
```

As before, the complement of every vector is also contained in the collection, and every two successive elements consists of one 0 and one 1. By stringing these vectors together to produce vectors $x(h) = (y(h), y(h), …)$, the resulting $x(h)$ vectors will include vectors produced for the previous level when h ranged from 1 to 4.

If it is desired to go farther, we may produce the pairs (y(p), y(q)) from this collection to produce 16 x 16 = 256 new y(h) vectors, each containing 8 + 8 = 16 components. These strongly balanced vectors do not possess some of the key features of vectors generated by the other methods described in this paper, and hence produce collections that are less diversified. Nevertheless, we anticipate that their novel structure may prove useful in certain types of applications.